\documentclass{article}
\usepackage{amsmath}
\usepackage{algorithm}
\usepackage{graphicx}
\usepackage{tcolorbox}
\usepackage{amsfonts}
\usepackage{algpseudocode}
\usepackage{framed}
\usepackage{mdframed}
\usepackage{url}
\usepackage{float}

\title{HyperArm Bandit Optimization: A Novel approach to Hyperparameter Optimization and an Analysis of Bandit Algorithms in Stochastic and Adversarial Settings}
\author{Samih Karroum \\ University of Ottawa \\ skarr062@uottawa.ca 
\and 
Saad Mazhar \\ University of Ottawa \\ smazh025@uottawa.ca}

\date{}

\begin{document}
\maketitle


\begin{abstract}
    This paper explores the application of bandit algorithms in both stochastic and adversarial settings, with a focus on theoretical analysis and practical applications.
    The study begins by introducing bandit problems, distinguishing between stochastic and adversarial variants, and examining key algorithms such as Explore-Then-Commit (ETC), Upper Confidence Bound (UCB), and Exponential-Weight Algorithm for Exploration and Exploitation (EXP3).
    Theoretical regret bounds are analyzed to compare the performance of these algorithms.
    The paper then introduces a novel framework, HyperArm Bandit Optimization (HABO), which applies EXP3 to hyperparameter tuning in machine learning models.
    Unlike traditional methods that treat entire configurations as arms, HABO treats individual hyperparameters as super-arms, and its potential configurations as sub-arms, enabling dynamic resource allocation and efficient exploration.
    Experimental results demonstrate HABO's effectiveness in classification and regression tasks, outperforming Bayesian Optimization in terms of computational efficiency and accuracy.
    The paper concludes with insights into the convergence guarantees of HABO and its potential for scalable and robust hyperparameter optimization.
\end{abstract}

\section{Introduction to Bandit Problems}
Bandit problems refer to when an agent must repeatedly choose between multiple options (commonly called arms) over a series of rounds. Each arm, when selected, provides the reward associated with that arm. The goal of a bandit is to choose actions that maximize the total accumulated reward over a given period of time, or number of rounds \textit{T}.\\

Within bandit problems, there exist two main variants, which this paper will cover. Firstly, \textit{stochastic} bandit problems are a variant in which the rewards for each arm are drawn from fixed but unknown probability distributions. Secondly, \textit{adversarial} bandit problems refer to when an adversary chooses the reward structure for every arm at each iteration, meaning the reward distribution for each arm is non-fixed. \\

The objective of this paper is to examine and compare different agents (in the form of algorithms) for both stochastic and adversarial bandits, and to look at their comparative performances via regret bounds analysis\cite{lattimore2020bandit, auer2002finite}. Furthermore, using the fundamental knowledge of bandits, the paper delves into their effectiveness in hyperparameter optimization.

\section{Introduction to ETC}
The Explore-Then-Commit (ETC) algorithm is an approach to the multi-armed stochastic bandit problem. It explores each arm (or option) \textit{m} number of times until it settles on the best possible reward.

\section*{Stochastic Bandits}

\subsection*{ETC (Explore Then Commit)}

Section 6.1-6.2

\paragraph{ETC} Plays each arm multiple times until a good estimate is achieved.

\paragraph{ETC Algorithm}
\[
\hat{\mu}_t(a) = \frac{1}{T_t(a)} \sum_{s=1}^{t} \mathbb{I}\{A_s = a\} X_s
\]
where \( T_t(a) \) is the number of times action \( a \) is played after round \( t \).

\paragraph{ETC Policy}
\[
A_t = 
\begin{cases}
(t \mod k) + 1 & \text{if } t \leq mk \\
\arg \max_a \hat{\mu}_t(a) & \text{if } t > mk
\end{cases}
\]
where \( k = \text{number of actions} \), and \( m = \text{minimum number of times each arm is explored} \).

\paragraph{Mean Reward}
\( \mu_a \) is the mean reward when the action \( a \) is played. 

\[
\Delta_a = \mu^* - \mu_a
\]
where \( \mu^* \) is the mean of the best action and \( \mu_a \) is the mean of action \( a \).

\subsection{ETC Simplified}
The steps the algorithm takes can be broken into two distinct phases:

\begin{enumerate}
    \item \textbf{Exploration:} Explore each arm a fixed number of times to learn about their respective rewards.
    \item \textbf{Commit:} Commit to the arm that has provided the best observed reward.
\end{enumerate}

\section{Algorithm Analysis}

Let us lay out the framework for the computation behind the algorithm:

Let $k$ represent the number of actions taken (arms).

Let $M$ represent the number of times each arm is explored.

Let $t$ represent the last round explored.

Let $Mk$ represent the number of rounds completed before choosing an action.

Let $i$ represent the arm from which the reward is received.

Let $\hat{\mu}_i(t)$ represent the average reward from arm $i$ after round $t$.

The summation is as follows:
    
\[
\hat{\mu}_i(t) = \frac{1}{T_i(t)} \sum_{s=1}^{t} \mathbb{I}\{A_s = i\} X_s,
\]

\label{fig:ETC summation}

Which can be written formally as the average reward from round 1 to $t$ of the indicator function
\(A_s=i\), which is an indicator function equal to 1 if the action \(A_s\) chosen at time \(s\) is equal to \(i\) and 0 otherwise.

We define \( X_s \) as the reward variable, where if the action \( A_s \) taken at time \( s \) is \( i \), the reward is \( X_s \); otherwise, it is zero. The cumulative reward is then calculated as the sum of these rewards divided by the total number of times that action \( i \) was chosen up to time \( t \).

However, this formula only represents an average reward. The policy of the ETC algorithm can be seen in the following.

\begin{framed}
    
\textbf{Algorithm 1: Explore-then-commit}

\textbf{Input:} $m$.

\textbf{In round} $t$, \textbf{choose action}
\[
A_t = 
\begin{cases}
(t \mod k) + 1, & \text{if } t \leq mk; \\
\arg\max_i \hat{\mu}_i(mk), & t > mk.
\end{cases}
\]

\label{fig:ETC policy}
\end{framed}

Then for round $t$, if $t \leq mk$, we take the modulus of the number of actions taken by round $t$ and add 1. This gives the remainder of $t$ of $k$, which is equal to $k - 1$. When we add 1, it ensures that the summation ranges from 1 to $k$ and that each arm is chosen cyclically.

\paragraph{Example}
\[
k=3 \text{ (3 arms)} \quad m=2 \quad \text{(2 explorations per arm)} \quad mk = 6 \text{ (total rounds)}
\]
\[
\begin{aligned}
t = 1, & \quad A_1 = (1 \mod 3) + 1 = 2 \\
t = 2, & \quad A_2 = (2 \mod 3) + 1 = 3 \\
t = 3, & \quad A_3 = (3 \mod 3) + 1 = 1 \\
t = 4, & \quad A_4 = (4 \mod 3) + 1 = 2 \\
t = 5, & \quad A_5 = (5 \mod 3) + 1 = 3 \\
t = 6, & \quad A_6 = (6 \mod 3) + 1 = 1 \\
\end{aligned}
\]

Here we see that we can use strong induction to deduce that the sequence of arms will continue to be \(2, 3, 1\) repeating.

\textbf{Figure 2.3.1: Explore-then-Commit (ETC) policy example}
\label{fig:ETC policy example}

The indicator function \(A_s = i\) is crucial for filtering out rewards not associated with the specific arm.
For instance, if $A_t = 2$ but $i = 3$, the rewards of $A_t$ should not be included in arm 3. 
This ensures accurate reward observations for all arms.
Upon conclusion of the final reward, the argmax (greatest observed reward) is repeatedly chosen.

\subsection{ETC Algorithm Explored Further}

\textbf{Step 1: Explore}

Calculate the reward of $A_s$ at action $i$, use the indicator function to ensure they are the same arm, and add $X_s$ if they are.

\[
\hat{\mu}_i(t) = \frac{1}{T_i(t)} \sum_{s=1}^{t} \mathbb{I}\{A_s = i\} X_s,
\]

\noindent\textbf{Step 2: Explore Part 2}

Ensure that each arm is run an equal number of rounds and sequentially before the conclusion of the final round. \\

\noindent\textbf{Step 3: Commit}
\[
\arg \max \hat{\mu}_i(mk), \quad t > mk.
\]
The algorithm identifies the action with the highest empirical mean. If actions have the same reward, actions are chosen arbitrarily. \\

\noindent\textbf{Step 4: Commit Part 2}

After $t > mk$ rounds,  the algorithm will repeatedly select action $i$ (action with the highest observed reward). \\

\noindent\textbf{ETC Example:}

Set $k=3$, $m=2$, thus $mk=6$

After 5 completed rounds, observed empirical means for each arm are as follows:

\[
\hat{\mu}_1 = 0.4, \quad \hat{\mu}_2 = 0.7, \quad \hat{\mu}_3 = 0.5
\]

\[
A_t = \arg \max_i \hat{\mu}_i(m \cdot k) = 0.7 \, (\hat{\mu}_2)
\]

Since arm 2 is chosen (as it has the highest observed reward):
\[
\sum_{t=1}^{\infty} \mathbb{I}\{A_t = \hat{\mu}_2\}
\]

\section{Theoretical ETC Regret Bound Analysis}

Theorem 1: When ETC is interacting with any 1 sub-Gaussian bandit (bandit where round variance is 1) and $1 \leq m \leq n/k$, then 

\centering
\[
R_n \leq m \sum_{i=1}^{k} \Delta_i + (n - mk) \sum_{i=1}^{k} \Delta_i \exp \left( -\frac{m \Delta_i^2}{4} \right)
\]

\raggedright
This computation represents the regret bound, which is the difference between the optimal reward and the reward chosen ($R_n$)\cite{lattimore2020bandit}.

\centering
\[
m \sum_{i=1}^{k} \Delta_i
\]

\raggedright
In this above equation showing the first summation term of the regret bound equation, $m$ represents the number of times each arm is played multiplied by the summation of $1$ to $k$ (the number of arms) of $\Delta_i$.
Where $\Delta_i$ represents the difference between $\mu(i)^*$ (the maximum obtainable reward) and $\mu(i)$. 

\centering
\[
(n - mk) \sum_{i=1}^{k} \Delta_i \exp \left( -\frac{m \Delta_i^2}{4} \right)
\]

\raggedright
The former is added to this computation.
This is the summation of $1$ to $k$ of the variation of the optimal reward minus every reward.
It is then multiplied by the exponential of $e$ to the following equation: the number of explorations multiplied by the square of $\Delta_i$ divided by $-4$.

\centering
\[
\exp \left( -\frac{m \Delta_i^2}{4} \right)
\]

\raggedright
This is an exponentially decaying multiplier that indicates how quickly the probability of guessing the optimal action decreases after each action taken.
We see that the more $m$ increases, the smaller the term will get, ultimately reducing the exploitation regret.

\section{ETC Conclusion}

Let us now consider an example to illustrate the concepts discussed.

\begin{framed}
    \centering
    
    $m = 2$, \quad $\Delta_1 = 3$ \\
    $k = 3$, \quad $\Delta_2 = 7$ \\
    $n = 10$, \quad $\Delta_3 = 2$ \\

    \[
    R_n \leq 2 \sum_{i=1}^{k} \Delta_i + c \sum_{i=1}^{k} \Delta_i \exp \left( -\frac{n \Delta_i^2}{4} \right)
    \]

    \[
    \begin{aligned}
    R_n &\leq 2(2+3+7) + c \Bigg( 2 \left( \exp \left( -\frac{3^2 \cdot 10}{4} \right) \right) \\
    &+ 3 \left( \exp \left( -\frac{3^2 \cdot 10}{4} \right) \right) \\
    &+ 7 \left( \exp \left( -\frac{7^2 \cdot 10}{4} \right) \right) \Bigg)
    \end{aligned}
    \]

    \[
    \leq 28 + 6(0.2706 + 0.000091492 + 1.027 \times 10^{-10})
    \]

    \[
    \leq 28 + 0.2706 
    R_n \leq 28.2706
    \]
\end{framed}

When it is all put together, the goal is to create a good balance between exploration and exploitation steps.
When that balance is found, we are able to achieve an excellent estimate for the optimal reward.

\section{Applied ETC Regret Analysis}

Continuing with the above example, once the optimal arm has been found, the algorithm works to maximize the number of times this arm is selected, subject to a bound, as seen in the following equations:

\subsection*{(1) Expected count during exploration}
\[
    \mathbb{E}[T_i(n)] = m + (n - mk)
\]

\subsection*{(2) Probability bound on arm selection}
\[
    \mathbb{P}(A_{mk+1} = i) \leq m + (n - mk)
\]

\subsection*{(3) Probability of optimal arm identification}
\[
    \mathbb{P} \left( \hat{\mu}_i(mk) \geq \max_{j \neq i} \hat{\mu}_j(mk) \right)
\]
 \\

Within these equations, the values represent the following:

\begin{itemize}
    \item $\mathbb{E}[T_i(n)]$: The expected number of times an arbitrary arm $i$ is selected during both the exploration and exploitation phases.
    \item $m$: The number of times each arm is played during the exploration phase.
    \item $k$: The number of arms.
    \item $n$: The total number of rounds across both exploration and exploitation phases.
    \item $\mathbb{P}(A_{mk+1} = i)$: The probability that arm $i$ is selected after the exploration phase.
\end{itemize}

\hspace{2em}This equation shows that the expected number of times arm $i$ is chosen is equal to the number of times each arm is played during exploration ($m$) plus the remaining rounds ($n - mk$) multiplied by the probability that arm $i$ is chosen during exploitation.

\hspace{2em}This expected value is bounded by a similar expression where the probability is replaced by the chance that the empirical mean reward of arm $i$ after exploration is greater than or equal to the maximum empirical mean reward of all other arms.

\hspace{2em}In essence, the equation tells us that the expected number of times arm $i$ is selected is less than the probability that the empirical mean of arm $i$ is greater than or equal to the highest empirical mean of all arms other than $i$.
This gives us a bound on the expected value of arm $i$ being chosen, acknowledging that $i$ may not be the optimal arm.

\hspace{2em}The reason the first probability $\mathbb{P}(A_{mk+1} = i)$ is lower than the probability involving the empirical means is due to the nature of the ETC algorithm, which is affected by factors such as a potentially suboptimal number of exploration rounds and noise in the reward data.

\vspace{2\baselineskip}

\textbf{For Example:}
\begin{itemize}
    \item \textbf{Arm $i$} has a true mean $\mu_i = 0.5$, but due to good luck during exploration, its empirical mean is $\hat{\mu}_i(mk) = 0.7$. \\
    \item \textbf{Arm $j$} has a true mean $\mu_j = 0.8$, but due to random fluctuations, its empirical mean is $\hat{\mu}_j(mk) = 0.65$.
\end{itemize}

In this case, the empirical mean of arm $i$ appears higher due to random noise or luck, even though arm $j$ is the better arm in the long run.

\hspace{2em}The right hand side is then bounded by the following equation that essentially states that the probability of the highest empirical mean of the most optimal arm that was found by the algorithm is greater than that of the arbitrary arm i, which reads as follows:

\small
\begin{align*}
\mathbb{P} \left( \hat{\mu}_i(mk) \geq \max_{j \neq i} \hat{\mu}_j(mk) \right) 
&\leq \mathbb{P} \left( \hat{\mu}_i(mk) \geq \hat{\mu}_1(mk) \right) \\
&= \mathbb{P} \left( \hat{\mu}_i(mk) - \mu_i - ( \hat{\mu}_1(mk) - \mu_1 ) \geq \Delta_i \right).
\end{align*}

\hspace{2em}This is equal to the probability that the \textbf{difference between the empirical mean of arm \( i \) and its true mean}, \textbf{minus} the difference between the optimal arm’s empirical mean and its true mean, is \textbf{greater than or equal to} the gap between the optimal true mean and the true mean of arm \( i \).
As more rounds are explored, the empirical means begin to converge to their true means.

\hspace{2em}As this happens, the difference in this inequality shrinks, making it less likely to hold with each additional round, leading to a \textbf{decreasing} probability that this event occurs.

\hspace{2em}In other words, the more rounds you explore, the more accurate the empirical means become, and the less likely it is that arm \( i \) will appear better than the optimal arm purely due to random chance.
Over time, the empirical means \( \hat{\mu}_i(mk) \) and \( \hat{\mu}_1(mk) \) get closer to their true means \( \mu_i \) and \( \mu_1 \), which means the differences in the inequality get smaller and smaller.

\section*{How This Helps the ETC Algorithm:}
\hspace{2em}In the context of the ETC (Explore-Then-Commit) algorithm, this behavior is crucial.
During the exploration phase, the algorithm is gathering information to estimate the empirical means of each arm.
The convergence of the empirical means to their true values helps ensure that by the time the algorithm commits to one arm for exploitation, it is very likely that the algorithm is selecting the optimal arm.

\hspace{2em}As the probability of arm \( i \)'s empirical mean being greater than the optimal arm's empirical mean due to randomness decreases, the algorithm becomes more confident that the arm with the highest empirical mean after exploration is indeed the optimal choice to commit to.
This reduces the chance of the algorithm committing to a suboptimal arm, ensuring better performance in the long run.

\section{Checking if Sub Gaussian}

\hspace{2em}Since bandit algorithms rely on empirical means to make decisions on what arms to choose, it is important that the rewards found with the algorithm don't deviate too far from the optimal reward.
This is done by making sure that the difference between the difference of arm $i$'s empirical mean and true mean and difference of the optimal arms empirical mean and true mean are \(\sqrt{\frac{2}{m}} \text{-subgaussian}\), or more formally:

\[
\hat{\mu}_i(mk) - \mu_i - \left( \hat{\mu}_1(mk) - \mu_1 \right) \text{ is } \sqrt{\frac{2}{m}} \text{-subgaussian}.
\]

\vspace{2\baselineskip}

We now reference Corollary 5.5 from Lattimore and Szepesvári's Bandit Algorithms\cite{lattimore2020bandit}:

\begin{framed}
\textbf{Corollary 5.5.} Assume that $X_i - \mu$ are independent, $\sigma$-subgaussian random variables. Then for any $\epsilon \geq 0$,
\[
\mathbb{P} (\hat{\mu} \geq \mu + \epsilon) \leq \exp \left( -\frac{n\epsilon^2}{2\sigma^2} \right)
\quad \text{and} \quad 
\mathbb{P} (\hat{\mu} \leq \mu - \epsilon) \leq \exp \left( -\frac{n\epsilon^2}{2\sigma^2} \right),
\]
where $\hat{\mu} = \frac{1}{n} \sum_{t=1}^{n} X_t$.

\textbf{Proof:} By Lemma 5.4, it holds that $\hat{\mu} - \mu = \sum_{i=1}^{n}(X_i - \mu)/n$ is $\sigma/\sqrt{n}$-subgaussian. Then apply Theorem 5.3.

Now, applying the sub-Gaussian bound to the difference between two sub-Gaussian variables:

\[
\mathbb{P} \left( (\hat{\mu}_i(mk) - \hat{\mu}_1(mk)) - (\mu_i - \mu_1) \geq \Delta_i \right) \leq \exp\left( -\frac{n\Delta_i^2}{2(2\sigma^2)} \right),
\]
where $\Delta_i = \mu_1 - \mu_i$.

\end{framed}

\hspace{2em}We set \(\sigma=1\) for simplicity and get our previous equation.
This inequality now assumes that our equation is sub-gaussian.

\hspace{2em}This corollary helps us understand the trade-off between exploration and exploitation. When $m$ is too small the probability of commitment to a suboptimal arm grows and inversely when $m$ is too large the policy explores for too long and the initial regret grows.

\hspace{2em}The textbook explores an example where $k=2$, with the first arm being optimal.
The bounds then become:

\begin{equation}
R_n \leq m\Delta + (n - 2m)\Delta \exp\left( -\frac{m\Delta^2}{4} \right) \leq m \Delta + n \Delta \exp\left( -\frac{m\Delta^2}{4} \right)
\end{equation}

\hspace{2em} Here we explore the upper bound, or worst case scenario for any given $n$ or $m$.
The initial term $m\Delta$ ($\Delta =$ the difference between the expected reward of a suboptimal arm and the optimal arm) represents the exploration phase where $m$ is the number of rounds explored, and we incur regret for every exploration where there is a difference between arms.
The second term represents the exploitation phase, where as $m$ grows the regret incurred decays, and the probability of a suboptimal arm being chosen decays exponentially.

\hspace{2em}The right-hand side of the inequality trumps in size because it views all rounds in both phases rather than exclusively the exploitation phase.
This right-hand side term then exponentially decays due to it being $\geq n-2m$, making it always larger than the left-hand side.
When $n$ is very large the term becomes essentially a rounding error due to its small size.

\hspace{2em}On the other hand $m$, or the number of exploration rounds per arm is chosen deterministically, we can also observe this value dynamically to minimize regret.
So to balance the exploration and exploitation we do the following:

\[
m = \max \left\{ 1, \left\lceil \frac{4}{\Delta^2} \log \left( \frac{n \Delta^2}{4} \right) \right\rceil \right\}
\]

When this approach is taken regret is bounded by the following:

\[
    R_n \leq \min \left\{ n \Delta, \Delta + \frac{4}{\Delta} \left( 1 + \max \left\{ 0, \log \left( \frac{n \Delta^2}{4} \right) \right\} \right) \right\}.
\]

\section{Introduction to UCB}
\hspace{2em}The UCB1 (Upper Confidence Bound) Algorithm that is based on the principle of \textbf{optimism in the face of uncertainty} which states to assume an optimistic outcome when the outcome is uncertain. This principle encourages increased exploration, and using observed data to adjust future choices.

\paragraph{UCB1 Algorithm}

\[
\text{UCB}_i = \hat{\mu}_i + \sqrt{\frac{2 \ln t}{N_i}}
\]

Where \(\hat{\mu}_i\) represents the estimated mean reward for action \(i\) based on previous trials. \\

The second term in the equation \(\sqrt{\frac{2 \ln t}{N_i}}\) is the confidence interval term. This term can be further broken down into:
\begin{itemize}
    \item \(t\): The number of total trials thus far.
    \item \(N_i\): The number of times action \(i\) has been chosen. \\
\end{itemize}

The confidence interval term ensures that actions with higher uncertainty values are more likely to be chosen, creating the basis for the UCB1 algorithm. \\

\subsection{UCB Simplified}
The steps the algorithm takes are as follows:

\begin{enumerate}
\item\textbf{Initialization:\\}
To start, initialize the estimated mean reward \(\hat{\mu}_i\) and count of how many times each choice has been taken \(N_i = 0\) for all choices \(i\).
\item\textbf{Exploration:\\}
Select each choice \(i\) once, and observe the reward, updating \(\hat{\mu}_i\) accordingly. Set \(N_i = 1\) for all \(i\) as all choices have been explored once.
\item\textbf{Main Loop:}
    \begin{enumerate} 
        \item\textbf{Calculate UCB}\\
        For each action \(i\) calculate the upper confidence bound for i using the formula \(\text{UCB}_i = \hat{\mu}_i + \sqrt{\frac{2 \ln t}{N_i}}\), where \(t\) is the current time step.
        \item\textbf{Select Optimal Choice}\\
        Select the action \(i^*\) that maximizes \(UCB_i\) from the formula \(i^* = argmax(UCB_i)\)
        \item\textbf{Observe and Update}\\
        Observe the reward \(r^*\) from the chosen action \(i^*\) and update the estimated mean reward 
        \(\hat{\mu}_{i*} = \frac{\hat{\mu}_{i*} \times N_{i*} + r_{i*}}{N_{i*} + 1}\). Lastly, increment \(N_i^*\) by 1.
    \end{enumerate}
\end{enumerate}

\section{Algorithm Analysis}

We now have a basic understanding of the UCB1 algorithm and it's variables.

\[
\text{UCB}_i = \hat{\mu}_i + \sqrt{\frac{2 \ln t}{N_i}}
\]

Let us further our knowledge by working through a practical example. \\

\textbf{Problem Statement:}
We are given a slot machine with \textbf{3 arms}, each with a unknown reward distribution, we will list these distributions to show the optimal decisions:
\begin{itemize}
    \item \textbf{Arm 1:} expected reward of 1.0.
    \item \textbf{Arm 2:} expected reward of 0.5.
    \item \textbf{Arm 3:} expected reward of 0.8.
\end{itemize}

These rewards are obviously not known to start the algorithm. \\

Now let us commence the algorithm following the aforementioned steps from section 8.1.

\begin{enumerate}
    \item \textbf{Round 1:}\\
    First we shall pull every arm once and document the reward.
    \begin{itemize}
        \item Pull arm 1, reward is 1.0, thus \(\hat{\mu}_1\) = 1.0, \(N_1 = 1\)
        \item Pull arm 2, reward is 0.5, thus \(\hat{\mu}_2\) = 0.5, \(N_2 = 1\).
        \item Pull arm 3, reward is 0.8, thus \(\hat{\mu}_3\) = 0.8, \(N_3 = 1\).
    \end{itemize}
    \item \textbf{Round 2:}\\
    Every arm has now been explored once, and we now calculate UCB for every arm to decide our next choice using the equation from \textit{Figure 9.1}.
    \begin{itemize}
        \item For arm 1: \(\hat{\mu}_1 = 1.0\), \(N_1 = 1\), \(t = 3\)
        \[\text{UCB}_1 = 1 + \sqrt{\frac{2 \ln 3}{1}} = 1 + 1.386 = 2.386\]
        \item For arm 2: \(\hat{\mu}_2 = 0.5\), \(N_2 = 1\), \(t = 3\)
        \[\text{UCB}_2 = 0.5 + \sqrt{\frac{2 \ln 3}{1}} = 0.5 + 1.386 = 1.886\]
        \item For arm 3: \(\hat{\mu}_3 = 0.8\), \(N_3 = 1\), \(t = 3\)
        \[\text{UCB}_3 = 0.8 + \sqrt{\frac{2 \ln 3}{1}} = 0.8 + 1.386 = 2.186\]
    \end{itemize}
    We can observe that \(\text{UCB}_1\) has the highest UCB of 2.386, so we pull arm 1 again.\\

\item \textbf{Round 3:}\\
We pull arm 1 again and observe the results after obtaining a reward of 1 again. Our estimated mean reward stays the same \(\hat{\mu}_1 = \frac{1 + 1}{2} = 1\), while the number of explorations of arm 1 increases \(N_1 = 2\). We know recalculate the UCB values for all three arms with these new values.

\begin{itemize}
        \item For arm 1: \(\hat{\mu}_1 = 1.0\), \(N_1 = 2\), \(t = 4\)
        \[\text{UCB}_1 = 1 + \sqrt{\frac{2 \ln 4}{2}} = 1 + 0.924 = 1.924\]
        \item For arm 2: \(\hat{\mu}_2 = 0.5\), \(N_2 = 1\), \(t = 4\)
        \[\text{UCB}_2 = 0.5 + \sqrt{\frac{2 \ln 4}{1}} = 0.5 + 1.553 = 2.053\]
        \item For arm 3: \(\hat{\mu}_3 = 0.8\), \(N_3 = 1\), \(t = 4\)
        \[\text{UCB}_3 = 0.8 + \sqrt{\frac{2 \ln 4}{1}} = 0.8 + 1.553 = 2.353\]
\end{itemize}

Now, arm 3 has the highest UCB value, thus will be the next chosen choice. This highlights the UCB1 algorithm's optimism principle, which is fundamental to the algorithm. This optimism principle encourages exploration and allows UCB1 to continually update its estimates, choosing the arm that has the best balance between exploration and exploitation. \\

\item \textbf{Repeat:}\\
This process of choosing the largest UCB value choice, observing the result, and recalculating UCB values will be repeated for the desired duration of the UCB algorithm.
    
\end{enumerate}

\textbf{Concluding Thoughts:}\\
\hspace{2em}After many rounds, the UCB1 algorithm will eventually converge to repeated selection of the arm with the highest estimated mean reward (arm 1 in above example).
However, the key principle of UCB1 is the fact that the algorithm will initially continue to explore suboptimal arms to ensure accurate estimations for all arms and premature commitment.

\section{Additional UCB Algorithms}

\hspace{2em}Throughout this chapter we have referred to the specific Upper Confidence Bound algorithm as UCB1, why is this? As you may have been able to guess, there are many different variants of the UCB algorithm. \\

\hspace{2em}One such variant is \(\text{UCB}(\delta)\). As we already have an understanding of UCB1 at this point, we will discuss the main differences between UCB1 and \(\text{UCB}(\delta)\).

\centering
\[
    \text{UCB}_i(t - 1, \delta) = 
    \begin{cases} 
        \infty & \text{if } T_i(t - 1) = 0 \\ 
        \hat{\mu}_i(t - 1) + \sqrt{\frac{2 \log(1 / \delta)}{T_i(t - 1)}} & \text{otherwise.} 
    \end{cases}
\] 
Figure 10.1: $UCB(\delta)$ Algorithm

\vspace{1em}

\raggedright
\hspace{2em}This formula can look quite intimidating, however fundamentally it is extremely similar to the UCB1 algorithm, with the addition of a confidence parameter: \(\delta\).
To illustrate this, we will once again break down the formula and it's variables.

\begin{itemize}
    \item \textbf{\(\text{UCB}_i(t - 1, \delta)\)}: The UCB score for arm \textit{i} at time \textit{t} with confidence level \(\delta\).
    \item \textbf{\(T_i(t)\)}: The number of times arm \textit{i} has been selected up to time \textit{t}.
    \item \textbf{\(\hat{\mu}_i(t)\)}: The estimated mean reward of arm \textit{i} based on the algorithm thus far.
    \item \textbf{\(\delta\)}: The confidence parameter (commonly 0.1 or 0.05) that adjusts the amount of exploration done by the algorithm.
\end{itemize}

\hspace{2em}With this understanding of the \(\text{UCB}(\delta)\) algorithm, we start to see that it is very familiar to UCB1. In both cases, each arm \textit{i} is selected once to begin, and then the arm with the largest calculated UCB value is chosen. \\

\hspace{2em}Evidently, this comes with one major difference, the confidence parameter $\delta$. The confidence parameter $\delta$ in the \(\text{UCB}(\delta)\) algorithm influences the amount of exploration that is done. This is because the exploration term \(\sqrt{\frac{2 \log(1 / \delta)}{T_i(t - 1)}}\) is inversely correlated to the size of \(\delta\).

\hspace{2em}Lower \(\delta\) values lead to \textbf{higher confidence bounds} (larger exploration term), encouraging more exploration, and less exploitation.
Conversely, higher \(\delta\) values lead to \textbf{lower confidence bounds} (smaller exploration term), encouraging less exploration and more exploitation.

\hspace{2em}This makes the \(\text{UCB}(\delta)\) algorithm a powerful approach for balancing exploration and exploitation with the power of the confidence parameter.

\section{Regret Analysis for UCB Algorithms}

\hspace{2em}The regret of a UCB algorithm refers to how much cumulative reward was lost due to choosing suboptimal choices over time.
Essentially, quantifying the difference between the algorithm's achieved reward in comparison to the maximum possible reward if the optimal choice was always selected.

\hspace{2em}This regret analysis between UCB1 and \(\text{UCB}(\delta)\) differ slightly as a result of the confidence parameter \(\delta\) introduced in the latter, however they are fundamentally similar.

\centering
\[
R(T) = O\left(\sum_{i:\Delta_i>0} \frac{\log T}{\Delta_i}\right)
\]

\vspace{1em}This equation represents the upper bound (as shown by the Big O) on the expected regret after \textit{T} rounds. We will break this equation down as follows:
\begin{itemize}
    \item \textbf{R(T):} The expected regret after T rounds.
    \item \(\Delta_i\): The suboptimality gap (the difference between the expected reward of the optimal arm and arm \textit{i}).
    \item \(\Sigma\): Showing that the sum is taken over all suboptimal arms.
    \item log\textit{T}: The natural logarithm of the time horizon.
\end{itemize}

\raggedright
From examining this upper bound, we can learn a few things about UCB1 as an algorithm.
Firstly, the regret grows logarithmically over time \textit{T}, showcasing the importance of time in achieving minimal regret.
Secondly, arms that are very suboptimal (large \(\Delta_i\)) contribute less to the regret, as they are less likely to be chosen during exploration.

\centering
\[
R(T) = O\left(\sum_{i:\Delta_i>0} \frac{\log(\log(T))}{\Delta_i}\right)
\]

\raggedright
\hspace{2em}When examining this equation we see that \(\text{UCB}(\delta)\)'s regret is extremely similar to UCB1's, except for one difference in the numerator, of the log(log(\textit{T})) term. While this may seem minor, this showcases a few key differences in the regret analysis of \(\text{UCB}(\delta)\). The change in the numerator means \(\text{UCB}(\delta)\) has a slower regret growth than UCB1. In practice, this means that \(\text{UCB}(\delta)\) has better asymptotic performance in theory, however careful selection of a good \(\delta\) is necessary for optimal performance. \\

\section{Closing Thoughts on UCB Algorithms} 

\hspace{2em}In conclusion, both UCB1 and \(\text{UCB}(\delta)\) demonstrate superior practical performance than ETC (by analysis of the regret bounds).
UCB1 achieves the optimal logarithmic regret shown above, which is a significant improvement on ETC's regret bounds.
\(\text{UCB}(\delta)\) further refines this with a double logarithmic regret, showing the advantages of an adaptive exploration strategy (with the use of the confidence parameter \(\delta\)).

\hspace{2em}This comparative regret analysis showcases how the UCB algorithms ability to balance exploration and exploitation (contrary to ETC'd rigid seperation of these phases) allows it to more efficiently identify and exploit optimal arms.

\newpage
\section*{EXP3}
\addcontentsline{toc}{section}{EXP3}

\section{Introduction to Adversarial Bandits}

\hspace{2em}The multi-armed bandit (MAB) problem is a classical framework for modeling decision-making under uncertainty, where an agent must choose among multiple options (or arms) to maximize cumulative rewards over time.
Traditional algorithms like Upper Confidence Bound (UCB) and Explore-Then-Commit (ETC) rely on stochastic assumptions about the reward distributions and use past statistical trends to make decisions.

\hspace{2em}However, in adversarial settings, the rewards can be assigned arbitrarily, potentially by an adversary aiming to minimize the agent's performance.
This necessitates algorithms that do not depend on stochastic properties of the rewards.
The \textit{EXP3} algorithm, introduced in 2001\cite{auer2001nonstochastic}, addresses this by ensuring exploration of every arm, regardless of past observations.

\section{EXP3 Algorithm}

\hspace{2em}EXP3 stands for \textit{Exponential-Weight Algorithm for Exploration and Exploitation}. It approaches the adversarial bandit problem by maintaining a probability distribution over the arms, ensuring continuous exploration while exploiting accumulated knowledge.

\subsection{Importance-Weighted Estimator}

\hspace{2em}In the MAB problem, only the reward of the selected arm is observed at each time step. To estimate the expected rewards for all arms, \textit{importance-weighted estimator} is used:

\[
\hat{X}_{ti} = \frac{\mathbb{I}\{A_t = i\} X_t}{P_{ti}},
\]

where: 
\begin{itemize} 
    \item $\mathbb{I}\{A_t = i\}$ is the indicator function, equal to $1$ if arm $i$ is selected at time $t$, and $0$ otherwise.
    \item $X_t$ is the observed reward at time $t$.
    \item $P_{ti}$ is the probability of selecting arm $i$ at time $t$.
\end{itemize}

This estimator ensures unbiased estimates of the expected rewards for each arm.

\section{Hedge Algorithm}
\hspace{2em}The \textit{Hedge Algorithm}\cite{freund1995hedge} is a fundamental method in online learning, used to combine expert advice by maintaining weights for each expert and updating them based on observed losses or rewards.
In the context of EXP3, we adapt the hedge algorithm to update cumulative rewards and derive selection probabilities.

\hspace{2em}Since we have an estimator, we begin to use what is known as the hedge algorithm, in which we give our values from the Importance-weighted estimator to the hedge algorithm. This is what ends up creating the EXP3 algorithm.

\hspace{2em}The Hedge algorithm itself is a full feedback algorithm, opposed to EXP3  which is a limited feedback algorithm.
The hedge algorithm being the foundation of EXP3 is adapted to fit the needs of a bandit or ``limited'' feedback algorithm.

\hspace{2em}The hedge algorithm operates by evaluating the loss, or ``feedback'', for all available arms at each round, rather than relying solely on the feedback from a single arm chosen in a given round.
This approach can be likened to a game where, at every move, the player has insight into the potential outcomes and rewards of all future options, enabling fully informed decision-making at each step.

Let us initialize some values, to further our understanding through an example.
\vspace{1em}

\textbf{Initial Values:}
\begin{itemize}
    \item Let $K$ be number of arms for every round $t$.
    \item Let $t$ be the round over $T$ rounds.
    \item Let $T$ be the total desired round.
    \item Let $w$ represent the weight of some arm.
    \item Let $\mathbb{P}(t, i) =$  the probability of arm $i$ being selected at round $t$.
    \item Let $\mathbb{W}(t, i) =$ the weight of arm $i$ at time $t$.
    \item Let $\mathbb{X}(t, i) =$ the reward of arm $i$ at time $t$.
\end{itemize}

\vspace{1em}With the starting values initialized, the general framework can be used as follows:

\begin{enumerate}
    \item Assign equal weights to each arm.
    \item Update weights based on loss incurred
    \item Pick arms based on higher weights and repeat
\end{enumerate}

The hedge algorithm balances exploration and exploitation, but will eventually converge to picking those with the highest weights.

\vspace{1em}

\textbf{Step 1:}

Diving into the Theory Further, we begin by initializing every weight for each arm, and every arm at the beginning has the same weight as follows, for example at \( t = 1 \), all weights are initialized equally:

\[
w_{1,i} = x \quad \forall i.
\]

\textbf{Step 2:}

We initialize the probabilities of each arm to begin, and selection of an arm is based on the relative weights of all of the arms.
The probability of choosing action \( i \) at time \( t \) is given by:

\[
p_{t,i} = \frac{w_{t,i}}{\sum_{j=1}^{K} w_{t,j}}.
\]

\vspace{1em}

\textbf{Step 3:}

After observing the probabilities, we follow the process of updating the weights, employing a weight-based update rule:

\[
    w_{t+1,i} = w_{t,i} \cdot \exp(\eta \cdot x_{t,i})
\]

\vspace{1em}
In this function, the weight of the arm in the next round changes based on its performance.
For example if $\mathbb{X}(t,i)$ is larger, then the weight of $w$ at $t+1$ will increase as this is an exponentially growing function.
This means in future rounds, those with high rewards will have higher weights, resulting in a higher probability of future selection.
The selection process is also influenced by the learning rate $\eta$.
Typically the following function is used in its definition:
\vspace{1em}

\[
\eta = \sqrt{\frac{\log(K)}{nK}}
\]

\vspace{1em}
The learning rate dynamically adjusts based on the number of rounds, typically starting with a larger value to encourage exploration.
As the rounds progress, the learning rate gradually decays, allowing the algorithm to converge towards selecting optimal choices.

\hspace{2em}This learning rate helps us achieve sub-linear regret, because as the number of rounds $n$ increases, the average regret per round will begin to approach zero, ensuring that the optimal choice is chosen.

\hspace{2em} $\log(K)$ is there to account for the number of arms and to balance exploration based on such.
$\log(K)$ will grow slowly to account for as many arms in $k$ as possible, promoting exploration.

\hspace{2em}The learning rate will always decrease overtime as more results are observed.
In the early rounds of the algorithm $\eta$ will be large and $n$ will be smaller, promoting exploration.
As $n$ grows larger, exploration is decreased for further exploitation.

\vspace{1em}

\textbf{Step 4:}

We calculate the total loss of each round by taking the sum of every arms reward multiplied by its probability at every round:

\[
\mathbb{E}[\ell_t] = \sum_{i=1}^{K} p_{t,i} \cdot \ell_{t,i},
\]

\vspace{1em}
This algorithm is the expected loss incurred by said round.
It helps us predict the expected loss to measure performance of the algorithm.
It gives us our average loss that we can expect at the end of the round.

\vspace{1em}

\textbf{Step 5:}

We get the cumulative loss of the hedge algorithm by calculating the sum of the expected loss at each round subtract the sum of the minimum loss possible, seen as follows:

\[
R_T = \sum_{t=1}^{T} \mathbb{E}[\ell_t] - \min_{i} \sum_{t=1}^{T} \ell_{t,i}.
\]

\vspace{1em}

This algorithm helps us see how the hedge algorithm performs against the optimal action.
We calculate the total regret and are able to see the difference in reward compared to if optimal action was constantly chosen.
This means, the lower the regret the better the algorithm.
The hedge algorithm's regret bounds as sub-linear which is important in bandit algorithms.

\section{Hedge Algorithm Applied}

Suppose we initialize the following values:

\begin{itemize}
    \item $K = 3 \text{ - Number of Arms}$
    \item $T = 2 \text{ - Number of Rounds}$
    \item $\eta_1 = 0.398 \text{ - Learning Rate for Round 1}$
    \item $\eta_2 = 0.282 \text{ - Learning Rate for Round 2}$
    \item $w_1 = w_2 = w_3 = 1$ \text{ - Initial Weights For Each Arm}
    \item $\ell_1 = 0.4$ \text{ - Loss for Arm 1}
    \item $\ell_2 = 0.3$ \text{ - Loss for Arm 2}
    \item $\ell_3 = 0.5$ \text{ - Loss for Arm 3}
\end{itemize}

\vspace{1em}

We can now calculate the probability of each arm's selection:

\[
p_{1,1} = \frac{w_{1,1}}{w_{1,1} + w_{1,2} + w_{1,3}} = \frac{1}{3}, \quad p_{1,2} = \frac{1}{3}, \quad p_{1,3} = \frac{1}{3}.
\]

\vspace{1em}
Next, we calculate the expected loss for this round.

\[
\mathbb{E}[\ell_1] = p_{1,1} \cdot \ell_{1,1} + p_{1,2} \cdot \ell_{1,2} + p_{1,3} \cdot \ell_{1,3} = \frac{1}{3} (0.4 + 0.3 + 0.5) = 0.4.
\]

\vspace{1em}
We can now begin the second round, following a similar process with the updated weights to round 1.

\vspace{1em}

\textbf{Step 1:}

Calculate new weights for round 2 using the updated learning rate \(\eta = 0.398\) for round 1:

    \[
    w_{2,1} = w_{1,1} \cdot e^{-\eta_1 \cdot \ell_{1,1}} = 1 \cdot e^{-0.398 \cdot 0.4} \approx 0.8521,
    \]
    \[
    w_{2,2} = w_{1,2} \cdot e^{-\eta_1 \cdot \ell_{1,2}} = 1 \cdot e^{-0.398 \cdot 0.3} \approx 0.8869,
    \]
    \[
    w_{2,3} = w_{1,3} \cdot e^{-\eta_1 \cdot \ell_{1,3}} = 1 \cdot e^{-0.398 \cdot 0.5} \approx 0.8190.
    \]

\vspace{1em}
\textbf{Step 2:}

Calculate new probabilities \( p_{2,i} \):

    \[
    p_{2,1} = \frac{w_{2,1}}{w_{2,1} + w_{2,2} + w_{2,3}} = \frac{0.8521}{0.8521 + 0.8869 + 0.8190} \approx 0.327,
    \]
    \[
    p_{2,2} = \frac{w_{2,2}}{0.8521 + 0.8869 + 0.8190} \approx 0.341,
    \]
    \[
    p_{2,3} = \frac{w_{2,3}}{0.8521 + 0.8869 + 0.8190} \approx 0.315.
    \]

\vspace{1em}
\textbf{Step 3:}
Calculate expected loss for round 2:

    \[
    \mathbb{E}[\ell_2] = p_{2,1} \cdot \ell_{2,1} + p_{2,2} \cdot \ell_{2,2} + p_{2,3} \cdot \ell_{2,3} = 0.327 \cdot 0.2 + 0.341 \cdot 0.4 + 0.315 \cdot 0.3.
    \]
    \[
    \mathbb{E}[\ell_2] \approx 0.0654 + 0.1364 + 0.0945 \approx 0.2963.
    \]

\vspace{1em}
\textbf{Step 4:}
Update weights for round 3 using \(\eta = 0.282\):

\[
w_{3,1} = w_{2,1} \cdot e^{-\eta_2 \cdot \ell_{2,1}} = 0.8521 \cdot e^{-0.282 \cdot 0.2} \approx 0.8057,
\]
\[
w_{3,2} = w_{2,2} \cdot e^{-\eta_2 \cdot \ell_{2,2}} = 0.8869 \cdot e^{-0.282 \cdot 0.4} \approx 0.8258,
\]
\[
w_{3,3} = w_{2,3} \cdot e^{-\eta_2 \cdot \ell_{2,3}} = 0.8190 \cdot e^{-0.282 \cdot 0.3} \approx 0.7925.
\]

\vspace{1em}
This completes the updated calculations for each step using \(\eta = 0.398\) for the first round and \(\eta = 0.282\) for the second round.

\section{Observations From Hedge Algorithm}

\hspace{2em}We are able to observe that after two rounds of the hedge algorithms, the weights for each arm were updated based on loss incurred, starting with an initial value of 1.
After implementing these changes, the expected loss for the current round decreased from 0.4 to 0.2963, representing a significant reduction of approximately 0.1.

\hspace{2em}While further rounds could potentially decrease the loss even more, the focus of this section is not on iterative improvements but rather on providing a foundational understanding of the theoretical basis of the EXP3 algorithm.
This explanation aims to help the reader grasp the underlying mechanisms of the algorithm.
Notably, the hedge algorithm, a full-information algorithm, updates the weights of all arms after each round.
In the following, we explore how the EXP3 algorithm adapts the principles of the hedge algorithm to address the challenges of the limited-feedback setting.

\section{EXP3 Algorithm}

EXP3 differs from the Hedge algorithm in its approach to decision-making across rounds.
While the Hedge algorithm updates the weights of all arms in each round, EXP3 selects a single arm based on a probability distribution and updates it accordingly.
This process leverages the previously introduced importance-weighted estimator, as detailed below.

\hspace{2em}The aforementioned hedge algorithm more closely resembles  the weight-based EXP3 algorithm\cite{auer2001nonstochastic}, however we will examine the reward based EXP3 algorithm\cite{lattimore2020bandit}, which is a derivation of the weight-based EXP3 algorithm.

\vspace{1em}
We initialize a set of starting values to proceed:

\begin{itemize}
    \item $K$ - Number of Actions
    \item $T$ - Number of rounds
    \item $\hat{S}_{t, i}$ - Cumulative Reward Received
\end{itemize}

\vspace{1em}
We initialize an learning parameter rate $\eta$ to provide a balance between exploration and exploitation. 
Depending on the size of $\eta$ we may try to aggressively exploit the best arm early on or further continue exploration.

\hspace{2em}We begin and initialize probability $P_{t,i}$ by taking the exponent of $\eta$ multiplied by the cumulative reward of the arm $i$ in the previous round divided by the sum of all available rewards.

\[
P_{t,i} = \frac{\exp\left( \eta \hat{S}_{t-1, i} \right)}{\sum_{j=1}^{k} \exp\left( \eta \hat{S}_{t-1, j} \right)}
\]

\vspace{1em}
An arm is then chose based on the previously derived probabilities, defined as action $A(t)$. 
The reward value $X(t,i)$ is then calculated of the chosen arm using the importance-weighed estimator to observe the reward gained in this round.

\[
\hat{X}_{t,i} = \frac{\mathbb{I}\{A_t = i\} \cdot X_t}{p_{t,i}}.
\]

\vspace{1em}
After this we update the cumulative reward based on our obtained results.

\[
\hat{S}_{t,i} = \hat{S}_{t-1, i} + 1 - \frac{\mathbb{I}\{A_t = i\} (1 - X_t)}{P_{t,i}}
\]

\vspace{1em}
We then move on to the next rounds and repeatedly update our reward values.

\section{EXP3 Algorithm Applied}

Let us initialize the following values:

\begin{itemize}
    \item $K=3$
    \item $X_t(1,1)=0.8$
    \item $X_t(2,1)=0.5$
    \item $X_t(3,1)=0.4$
    \item $T = 2$
    \item $\eta$ remaining the same
\end{itemize}

We also initialize the cumulative reward of all arms to 0 to begin.
\[
\hat{S}_{1,1} = \hat{S}_{1,2} = \hat{S}_{1,3} = 0
\]

\subsection{Round 1}

We begin by initializing our learning rate $\eta$:

\vspace{1em}
\[
\eta_1 = \sqrt{\frac{\log(3)}{1 \cdot 3}} \approx 0.398
\]

\vspace{1em}
Next, probabilities are calculated based on the initial rewards:

\vspace{1em}
\[
P_{1,i} = \frac{\exp(\eta_1 \cdot \hat{S}_{1,i})}{\sum_{j=1}^{K} \exp(\eta_1 \cdot \hat{S}_{1,j})} = \frac{\exp(0)}{3 \cdot \exp(0)} = \frac{1}{3} = 0.333
\]

\vspace{1em}

As all arms currently have an equal probability of selection ($\frac{1}{3}$), we arbitrarily choose an arm, say 2.
We then calculate arm 2's importance weighted estimator:

\[
\hat{S}_{1,i} = \hat{S}_{0, i} + 1 - \frac{\mathbb{I}\{A_1 = i\} (1 - X_1)}{P_{1,i}} = 0 + 1 - \frac{1 \cdot (1 - 0.5)}{0.333} \approx 1 - \frac{0.5}{0.333} \approx 1 - 1.5 = -0.5
\]

\vspace{1em}
We can now update the reward for action 2:

\[
\hat{S}_{1,2} = \hat{S}_{0,2} + 1 - \frac{\mathbb{I}\{A_1 = 2\} (1 - X_1)}{P_{1,2}} = 0 + 1 - \frac{1 \cdot (1 - 0.5)}{0.333} \approx 1 - 1.5 = -0.5
\]

\subsection{Round 2}

For the second round, the learning rate is recalculated as follows:

\vspace{1em}
\[
\eta_2 = \sqrt{\frac{\log(3)}{2 \cdot 3}} \approx 0.28
\]

\vspace{1em}
With the updated cumulative rewards from round 1, we calculate the probabilities for each arm based on \(\hat{S}_{2,i}\).

\vspace{1em}

\textbf{Arm 1:}
\[
P_{2,1} = \frac{\exp(0.28*0)}{\exp(0*0.28) + \exp(-0.14) + \exp(0*0.28)} \approx 0.349
\]

\textbf{Arm 2:}
\[
P_{2,2} = \frac{\exp(0.28 \cdot (-0.5))}{\exp(0.28*0) + \exp(-0.14) + \exp(0.28*0)} = \frac{\exp(-0.14)}{1 + \exp(-0.14) + 1} \approx 0.303
\]

\textbf{Arm 3:}
\[
P_{2,3} = \frac{\exp(0.28*0)}{1 + \exp(-0.14) + 1} \approx 0.349
\]

\vspace{1em}
Based on these probabilities, we arbitrarily choose arm 1.
We then observe the reward for action 1 as \(X_{2,1} = 0.8\) and calculate the importance-weighted estimator for the chosen action.

\textbf{Calculate Importance-Weighted Estimated Reward for Action 1:}
\[
\hat{S}_{2,1} = \hat{S}_{0, i} + 1 - \frac{\mathbb{I}\{A_1 = i\} (1 - X_1)}{P_{1,i}} = 0 + 1 - \frac{1 \cdot (1 - 0.8)}{0.349}\approx 1 - 0.573 = 0.427
\]

Since action 1 was the only one chosen, we update the cumulative rewards accordingly:

\[
\hat{S}_{3,1} = \hat{S}_{2,1} + \hat{X}_{2,1} = 0 + 0.427 = 0.427
\]

\subsection{Round 3: Calculate the Dynamic Learning Rate}

For the third round, we update the learning rate as follows:

\[
\eta_3 = \sqrt{\frac{\log(3)}{3 \cdot 3}} \approx 0.23
\]

With the updated cumulative rewards from round 2, we calculate the probabilities for each arm based on \(\hat{S}_{3,i}\):

\vspace{1em}
\textbf{Arm 1:}
\[
P_{3,1} = \frac{\exp(0.23 \cdot 0.427)}{\exp(0.23 \cdot 0.427) + \exp(0.23 \cdot (-0.5)) + \exp(0)} \approx  0.368
\]

\textbf{Arm 2:}
\[
P_{3,2} = \frac{\exp(0.23 \cdot (-0.5))}{\exp(0.23 \cdot 0.427) + \exp(0.23 \cdot (-0.5)) + \exp(0)} \approx 0.298
\]

\textbf{Arm 3:}
\[
P_{3,3} = \frac{1}{\exp(0.23 \cdot 0.427) + \exp(0.23 \cdot (-0.5)) + \exp(0)} \approx 0.33
\]

\subsection*{Observations from the EXP3 Example}

\hspace{2em}After two iterations of running the EXP3 algorithm, we initialized with equal weights and observed rewards solely for the chosen actions.
The weights were then updated using importance-weighted rewards, leading to an adjustment in probabilities that favored actions associated with higher rewards.
Unlike the Hedge algorithm, which updates all weights based on complete feedback, EXP3 adapts to limited feedback by relying on estimated rewards for the actions that were selected.
This demonstrates the algorithm’s ability to operate effectively under conditions of partial information.

\hspace{2em}In this example, after just two rounds, EXP3 has learned to favor the actions with observed rewards, illustrating how it balances exploration and exploitation by keeping some probability for each action through the \( \gamma \) parameter. This ensures that actions with potentially high rewards aren't ignored, even if they initially have lower weights. 

\hspace{2em}This setup shows the foundational adaptation of the Hedge algorithm to the limited-feedback setting in the EXP3 algorithm.

\section{Theoretical Foundations, Regret Analysis, and bounding proofs}

\subsection{Overview and Intuition}

The EXP3 algorithm operates by assigning probabilities to each arm by the exponential of their importance-weighted cumulative rewards. We present a proof defining the expected regret of the algorithm with respect to the best possible arm is on the order of \(\sqrt{K T \log(K)}\). \(T\) being the number of rounds, \(K\) being the number of arms. We introduce a weighted quantity \(W_t\) accounting for cumulative rewards at round t, and perform algebraic operations to bound its regret.

\subsection{Setting and Notation}

- We have \(k\) arms and run the algorithm for \(T\) rounds.
- Let \(x_{t,i} \in [0,1]\) be the \emph{reward} of arm \(i\) at time \(t\). Rewards are bounded in \([0,1]\).
- At each round \(t\), the algorithm chooses an arm \(A_t\) according to probabilities \(P_{t,i}\).
- To handle the partial feedback (only observing the chosen arm's reward), we define the importance-weighted estimator:
\[
\hat{X}_{t,i} = \frac{\mathbb{I}\{A_t = i\} x_{t,i}}{P_{t,i}},
\]
ensuring that \(\mathbb{E}_{t-1}[\hat{X}_{t,i}] = x_{t,i}\). This unbiasedness is crucial in our analysis.

We let:
\[
\hat{S}_{T,i} = \sum_{t=1}^{T} \hat{X}_{t,i}
\]
be the cumulative estimated reward for arm \(i\). Similarly, the total expected reward of the algorithm is:
\[
\sum_{t=1}^T \mathbb{E}_{t-1}[X_t] = \sum_{t=1}^T \sum_{i=1}^k P_{t,i} x_{t,i}.
\]
Since \(\hat{X}_{t,i}\) is an unbiased estimator, we have:
\[
\mathbb{E}[\hat{S}_{T,i}] = \sum_{t=1}^T x_{t,i}.
\]

For the algorithm's achieved cumulative reward, we consider:
\[
\hat{S}_T = \sum_{t=1}^T \sum_{i=1}^k P_{t,i}\hat{X}_{t,i}.
\]
Its expectation equals the sum of expected rewards chosen by the algorithm over time.

\subsection{Defining Regret}

We define the expected regret relative to a fixed arm \(i\):
\[
R_{T,i} = \sum_{t=1}^T x_{t,i} - \mathbb{E}\left[\sum_{t=1}^T X_t \right].
\]

Using the unbiased estimators, we rewrite this as:
\[
R_{T,i} = \mathbb{E}[\hat{S}_{T,i}] - \mathbb{E}\left[\sum_{t=1}^T \sum_{i=1}^k P_{t,i} \hat{X}_{t,i}\right] = \mathbb{E}[\hat{S}_{T,i} - \hat{S}_T].
\]

Intuitively, \(R_{T,i}\) measures how much better we could have done by always playing arm \(i\) versus what the algorithm actually achieved.

\subsection{Introducing \(W_t\) and Its Properties}

To prove a bound on regret, we introduce the key auxiliary quantity:
\[
W_t = \sum_{j=1}^k \exp(\eta \hat{S}_{t,j}),
\]
where \(\eta > 0\) is a learning rate to be chosen later. Initially, at round \(t=0\), we have no cumulative rewards:
\[
W_0 = \sum_{j=1}^k \exp(\eta \cdot 0) = \sum_{j=1}^k 1 = k.
\]

The quantity \(W_t\) can be seen as a normalization factor in the probabilities used by Exp3. The probabilities \(P_{t,i}\) are proportional to \(\exp(\eta \hat{S}_{t-1,i})\), ensuring that arms with higher estimated rewards become more likely to be chosen over time.

We observe that at any time:
\[
\frac{W_t}{W_{t-1}} = \frac{\sum_{j=1}^k \exp(\eta \hat{S}_{t,j})}{\sum_{j=1}^k \exp(\eta \hat{S}_{t-1,j})}.
\]

By factoring out \(\exp(\eta \hat{S}_{t-1,j})\) from the numerator and using the definition of \(P_{t,j}\) (which is proportional to these exponentials), we get:
\[
\frac{W_t}{W_{t-1}} = \sum_{j=1}^k \frac{\exp(\eta \hat{S}_{t-1,j})}{W_{t-1}}\exp(\eta \hat{X}_{t,j}) = \sum_{j=1}^k P_{t,j}\exp(\eta \hat{X}_{t,j}).
\]

\subsection{Bounding the Growth of \(W_t\)}

We now seek to bound \(\frac{W_t}{W_{t-1}}\) from above. Since \(x_{t,i}\in[0,1]\), we have \(\hat{X}_{t,j} \in [0, \frac{1}{P_{t,j}}]\). In particular, \(\hat{X}_{t,j}\) is bounded above by a quantity typically no larger than some constant (in expectation). Consider the inequality for \(x\leq 1\):
\[
\exp(x) \leq 1 + x + x^2.
\]
Also, for all real \(x\):
\[
1 + x \leq \exp(x).
\]

Apply the first inequality with \(x = \eta \hat{X}_{t,j}\) (assuming \(\eta \hat{X}_{t,j}\leq 1\), which we handle by choosing \(\eta\) appropriately):
\[
\exp(\eta \hat{X}_{t,j}) \leq 1 + \eta \hat{X}_{t,j} + \eta^2 \hat{X}_{t,j}^2.
\]

Substituting this into our expression for \(\frac{W_t}{W_{t-1}}\):
\[
\frac{W_t}{W_{t-1}} = \sum_{j=1}^k P_{t,j}\exp(\eta \hat{X}_{t,j}) \leq \sum_{j=1}^k P_{t,j}(1 + \eta \hat{X}_{t,j} + \eta^2 \hat{X}_{t,j}^2).
\]

Since \(\sum_{j=1}^k P_{t,j}=1\), this simplifies to:
\[
\frac{W_t}{W_{t-1}} \leq 1 + \eta \sum_{j=1}^k P_{t,j}\hat{X}_{t,j} + \eta^2 \sum_{j=1}^k P_{t,j}\hat{X}_{t,j}^2.
\]

Using \(1+x \leq \exp(x)\) on the entire right-hand side (and noting the argument is now a sum involving \(\hat{X}_{t,j}\)):
\[
\frac{W_t}{W_{t-1}} \leq \exp\left(\eta \sum_{j=1}^k P_{t,j}\hat{X}_{t,j} + \eta^2 \sum_{j=1}^k P_{t,j}\hat{X}_{t,j}^2\right).
\]

To extend this to all rounds \(t = 1, \dots, T\), we take the product over \(t\):
\[
\prod_{t=1}^T \frac{W_t}{W_{t-1}} \leq \prod_{t=1}^T \exp\left(\eta \sum_{j=1}^k P_{t,j}\hat{X}_{t,j} \;+\; \eta^2 \sum_{j=1}^k P_{t,j}\hat{X}_{t,j}^2\right).
\]

On the left-hand side, note that the product telescopes:

\[
\prod_{t=1}^T \frac{W_t}{W_{t-1}} = \frac{W_1}{W_0} \cdot \frac{W_2}{W_1} \cdot \frac{W_3}{W_2} \cdots \frac{W_T}{W_{T-1}} = \frac{W_T}{W_0}.
\]

Since \(W_0 = k\), we have:
\[
\frac{W_T}{k} \leq \prod_{t=1}^T \exp\left(\eta \sum_{j=1}^k P_{t,j}\hat{X}_{t,j} \;+\; \eta^2 \sum_{j=1}^k P_{t,j}\hat{X}_{t,j}^2\right).
\]

We now use the property that the product of exponentials is the exponential of the sum of the exponents. Specifically, if we set:
\[
f_t = \eta \sum_{j=1}^k P_{t,j}\hat{X}_{t,j} \;+\; \eta^2 \sum_{j=1}^k P_{t,j}\hat{X}_{t,j}^2,
\]
then:
\[
\prod_{t=1}^T \exp(f_t) = \exp\left(\sum_{t=1}^T f_t\right).
\]

Applying this:
\[
\frac{W_T}{k} \leq \exp\left(\sum_{t=1}^T \left(\eta \sum_{j=1}^k P_{t,j}\hat{X}_{t,j} \;+\; \eta^2 \sum_{j=1}^k P_{t,j}\hat{X}_{t,j}^2\right)\right).
\]

We can rewrite the sum inside the exponent by grouping terms:
\[
\sum_{t=1}^T \left(\eta \sum_{j=1}^k P_{t,j}\hat{X}_{t,j}\right) + \sum_{t=1}^T \left(\eta^2 \sum_{j=1}^k P_{t,j}\hat{X}_{t,j}^2\right) 
= \eta \sum_{t=1}^T \sum_{j=1}^k P_{t,j}\hat{X}_{t,j} \;+\; \eta^2 \sum_{t=1}^T \sum_{j=1}^k P_{t,j}\hat{X}_{t,j}^2.
\]

Multiplying both sides by \(k\):
\[
W_T \leq k \exp\left(\eta \hat{S}_T + \eta^2 \sum_{t=1}^T \sum_{j=1}^k P_{t,j}\hat{X}_{t,j}^2\right).
\]

Recall that \(\hat{S}_T = \sum_{t=1}^T \sum_{j=1}^k P_{t,j}\hat{X}_{t,j}\). Thus:
\[
W_T \leq k \exp\left(\eta \hat{S}_T + \eta^2 \sum_{t=1}^T \sum_{j=1}^k P_{t,j}\hat{X}_{t,j}^2\right).
\]

At the same time, by the definition of \(W_T\), for any fixed arm \(i\):
\[
\exp(\eta \hat{S}_{T,i}) \leq W_T.
\]

Combining these two inequalities:
\[
\exp(\eta \hat{S}_{T,i}) \leq k \exp\left(\eta \hat{S}_T + \eta^2 \sum_{t=1}^T \sum_{j=1}^k P_{t,j}\hat{X}_{t,j}^2\right).
\]

Taking the natural logarithm:
\[
\eta \hat{S}_{T,i} \leq \log(k) + \eta \hat{S}_T + \eta^2 \sum_{t=1}^T \sum_{j=1}^k P_{t,j}\hat{X}_{t,j}^2.
\]

Divide through by \(\eta>0\):
\[
\hat{S}_{T,i} - \hat{S}_T \leq \frac{\log(k)}{\eta} + \eta \sum_{t=1}^T \sum_{j=1}^k P_{t,j}\hat{X}_{t,j}^2.
\]

\subsection{Bounding the Second Moment Term}

We now bound the term involving $\hat{X}_{t,j}^2$. Recall:
\[
\hat{X}_{t,j} = \frac{\mathbb{I}\{A_t = j\} x_{t,j}}{P_{t,j}},
\]
and let $y_{t,j} = 1 - x_{t,j}$. Note that $0 \leq y_{t,j} \leq 1$ since $x_{t,j}\in[0,1]$.

Then:
\[
\hat{X}_{t,j} = 1 - \frac{\mathbb{I}\{A_t = j\} y_{t,j}}{P_{t,j}}.
\]
Squaring this:
\[
\hat{X}_{t,j}^2 = \left(1 - \frac{\mathbb{I}\{A_t = j\} y_{t,j}}{P_{t,j}}\right)^2 = 1 - 2 \frac{\mathbb{I}\{A_t = j\} y_{t,j}}{P_{t,j}} + \frac{\mathbb{I}\{A_t = j\} y_{t,j}^2}{P_{t,j}^2}.
\]

Taking the expectation and multiplying by $P_{t,j}$:
\[
\mathbb{E}\left[\sum_{j=1}^k P_{t,j}\hat{X}_{t,j}^2\right] = \mathbb{E}\left[\sum_{j=1}^k P_{t,j}\left(1 - 2\frac{\mathbb{I}\{A_t = j\}y_{t,j}}{P_{t,j}} + \frac{\mathbb{I}\{A_t = j\}y_{t,j}^2}{P_{t,j}^2}\right)\right].
\]

Distribute $P_{t,j}$:
\[
= \mathbb{E}\left[\sum_{j=1}^k P_{t,j} - 2\mathbb{I}\{A_t = j\}y_{t,j} + \frac{\mathbb{I}\{A_t = j\}y_{t,j}^2}{P_{t,j}}\right].
\]

Since $\sum_{j=1}^k P_{t,j} = 1$, and $\mathbb{E}[\mathbb{I}\{A_t=j\}y_{t,j}] = P_{t,j}y_{t,j}$, and using $y_{t,j}^2 \leq 1$, we find:
\[
\mathbb{E}\left[\sum_{j=1}^k P_{t,j}\hat{X}_{t,j}^2\right] \leq \mathbb{E}[1 - 2Y_t + k],
\]
where $Y_t = 1 - X_t$ and $X_t$ is the actual reward chosen at time $t$. Since $y_{t,j} \leq 1$, a more careful accounting shows this term is bounded by a constant, and standard results (from the literature on Exp3) indicate that:
\[
\mathbb{E}\left[\sum_{j=1}^k P_{t,j}\hat{X}_{t,j}^2\right] \leq k.
\]

Thus:
\[
\mathbb{E}[\hat{S}_{T,i} - \hat{S}_T] \leq \frac{\log(k)}{\eta} + \eta T k.
\]

\subsection{Substituting in $\eta$}

We substitute back into the equation the value of $\eta$, which is:
\[
\eta = \sqrt{\frac{\log(k)}{T k}}.
\]

Substitute this back into the bound:
\[
\mathbb{E}[\hat{S}_{T,i} - \hat{S}_T] \leq \frac{\log(k)}{\sqrt{\frac{\log(k)}{T k}}} + \sqrt{\frac{\log(k)}{T k}} T k = 2\sqrt{T k \log(k)}.
\]

\subsection{Conclusion}

Thus, we have shown:
\[
\mathbb{E}[R_{T,i}] = \mathbb{E}[\hat{S}_{T,i} - \hat{S}_T] \leq 2\sqrt{T k \log(k)}.
\]

This matches the known regret bound for Exp3, demonstrating that its performance is near-optimal in the adversarial bandit setting.

\section{Novel approach to hyperparameter tuning in machine learning models with Multi-Armed Bandit algorithms}
In machine learning models, hyperparameters are different configurations of a model used, where tuning them can result in much greater accuracy and prediction in machine learning tasks.
Hyperparameter tuning had been use case for bandit algorithms like in the case of the hyperband algorithm\cite{li2018hyperband}. The hyperband algorithm uses a bandit like framework, treating entire hyperparameter configurations as arms entirely, weeding out different configurations to converge to the best choices. Our research used a new and different approach to using MAB's to find the best hyperparameter configurations. Rather than treat different configurations as arms, hyperparameters themselves were treated as arms.

\section{Introduction to HABO}

\subsection{What is HABO?}
HyperArm Bandit Optimization (HABO) is a novel framework designed to efficiently handle hyperparameter optimization using multi-armed bandit (MAB) algorithms. Unlike traditional methods that consider full configurations as individual arms, HABO treats each hyperparameter as an independent super-arm, with its potential configurations as sub-arms. This innovative approach allows dynamic allocation of computational resources, balancing \textbf{exploration} (testing new configurations) and \textbf{exploitation} (refining promising configurations) based on performance feedback.

\subsection{Why HABO?}
Hyperparameter optimization is essential for improving the accuracy and efficiency of machine learning models, but many of the traditional methods have significant limitations:

\begin{itemize}
    \item \textbf{Grid Search}: Explores the entire hyperparameter space exhaustively, leading to a \textit{combinatorial explosion} as the number of parameters increases. Computationally expensive, especially for models with many hyperparameters or large ranges.
    \item \textbf{Random Search}: Stochastic nature often fails to focus on optimal regions of the search space. Does not leverage feedback to optimize its search, resulting in inefficient resource utilization.
    \item \textbf{Bayesian Optimization (BO)}: Relies on Gaussian Process (GP) models to predict rewards and guide search. Computationally expensive, with costs increasing significantly as dimensionality grows. Assumes smoothness in the optimization landscape, which limits its applicability in noisy or adversarial-like conditions.
\end{itemize}

HABO addresses these issues through:
\begin{itemize}
    \item \textbf{Dynamic Resource Allocation}: Efficiently focuses computational efforts on promising regions.
    \item \textbf{Scalability}: Treating hyperparameters independently reduces complexity.
    \item \textbf{Robustness}: Leveraging adversarial bandit algorithms like EXP3 ensures resilience to noise and irregularities in the optimization landscape.
\end{itemize}

\section{Methodology}

\subsection{Core Idea}
HABO reformulates hyperparameter optimization into a multi-armed bandit framework:
\begin{itemize}
    \item \textbf{Hyperparameters as Super-Arms}: Each hyperparameter is treated as a super-arm, or an arm that contains other arms/potential values. So for example in a random forests model, {n\_estimators} would be an arm, as would {max\_features}, {min\_samples\_split} and {max\_depth}.
    \item \textbf{Configurations as Sub-Arms}: The sub-arms represent all the potential options/configurations of the super arms. This would all be dependant on the users choice of a range, but essentially they represent the actual potential value of the super arm.
    \begin{itemize}
        \item \textit{Numerical Parameters}: Discretized into ranges with a defined step size chosen by user (e.g., \texttt Random forests {n\_estimators} as $[50, 60, \dots, 300]$), $\min=50$, $\max=300$, step $=10$.
        \item \textit{Categorical Parameters}: Mapped directly to discrete choices (e.g., \texttt Random Forests {max\_features} as $[\text{sqrt}, \text{log2}, \text{None}]$).
    \begin{figure}[htbp]
        \centering
        \includegraphics[width=0.75\linewidth]{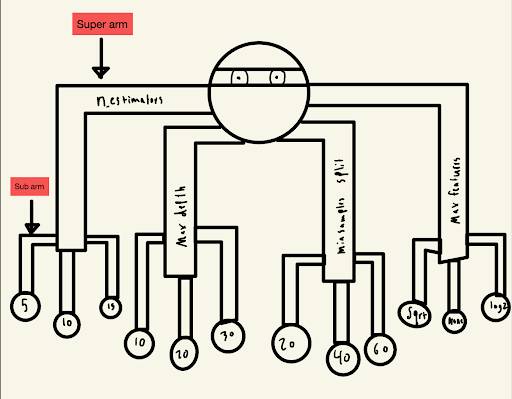}
        \caption{Visual representation of HABO\cite{karroum2024github}}
        \label{fig:enter-label}
    \end{figure}
    \end{itemize}
\end{itemize}

\subsection{Algorithm Overview}
HABO employs the \textbf{EXP3 algorithm}, but not using the previously discussed Value-Based version for super-arms.  In the implementation the weight-update variant of EXP3 was used rather than the value-based approach for the sub-arm updates.
This implementation introduces a gamma variable, an adjustable exploration setting dependent on the users needs. This is important as with a potential huge number of sub arms, converging to one value that may be optimal in retrospect might prove inefficient, diverging from a potential better configuration.
\begin{framed}
\begin{algorithm}[H]
\caption{EXP3 Algorithm (Weight-Based)}
\textbf{Input:} $T$ (number of iterations), $K$ (number of arms), $\gamma$ (exploration parameter)\\
\textbf{Initialization:} Set $w_i = 1$ for all $i \in \{1, \dots, K\}$ (initialize weights equally)
\begin{algorithmic}[1]
\For{$t = 1, \dots, T$}
    \State Compute the sampling distribution:
    \[
    P_{t,i} = (1 - \gamma) \frac{w_i}{\sum_{j=1}^K w_j} + \frac{\gamma}{K}, \quad \forall i \in \{1, \dots, K\}
    \]
    \State Sample an arm $i$ according to $P_t$ and observe the reward $X_t$
    \State Update the weight of the chosen arm:
    \[
    w_{i} \leftarrow w_{i} \cdot \exp\left(\frac{\gamma X_t}{K P_{t, i}}\right)
    \]
\EndFor
\end{algorithmic}
\end{algorithm}
\end{framed}

\textit{The first iteration of the EXP3 algorithm drawn from the work of Auer\cite{auer2001nonstochastic}}

\subsection{Dynamic Resource Allocation}
The HABO framework follows these steps:
\begin{enumerate}
    \item \textbf{Select Hyperparameter}: Use a bandit strategy (e.g., EXP3) to pick a hyperparameter.
    \item \textbf{Explore Configurations}: Sample sub-arms within the chosen hyperparameter based on calculated probabilities.
    \item \textbf{Allocate Resources}: Assign more computational effort to configurations with better rewards, progressively refining the search space.
\end{enumerate}

\subsection{Hyper-Arm Bandit Optimization Pseudocode}

\begin{algorithm}[H]
\caption{HyperArm Bandit Optimization: Exp3 Implementation}
\textbf{Input:} $T$ (number of iterations), $k$ (number of hyperparameters), $n_i$ (number of possible values for each hyperparameter), $\gamma$ (exploration parameter) \\
\textbf{Initialization:} Initialize weights $\{w_{i,j}\}$ to $1$ for all $i \in \{1, \dots, k\}$ and $j \in \{1, \dots, n_i\}$

\begin{algorithmic}[1]
\For{$t = 1, \dots, T$}
    \State \textbf{Select a hyperparameter to adjust:}
    \begin{itemize}
        \item Compute the probability distribution over super-arms:
        \[
        P_{t,i} = (1-\gamma) \frac{w_i}{\sum_{j=1}^k w_j} + \frac{\gamma}{k}
        \]
        \item Sample a super-arm $I_t \sim P_t$ and observe its configuration space. 
    \end{itemize}

    \State \textbf{Select a value for the chosen hyperparameter:}
    \begin{itemize}
        \item Compute the probability distribution over sub-arms for the selected super-arm $I_t$:
        \[
        Q_{t,j|I_t} = \frac{w_{I_t, j}}{\sum_{j=1}^{n_{I_t}} w_{I_t, j}}
        \]
        \item Sample a sub-arm $J_t$ within super-arm $I_t$ and set the hyperparameter value accordingly.
    \end{itemize}

    \State \textbf{Train the model:}
    \begin{itemize}
        \item Update the hyperparameter configuration with $J_t$
        \item Train the model using the current hyperparameter configuration
        \item Evaluate the model and compute the reward $R_t$:
        \begin{itemize}
            \item \textbf{Classification:} $R_t = \text{Accuracy}$
            \item \textbf{Regression:} $R_t = R^2$
        \end{itemize}
    \end{itemize}

    \State \textbf{Update weights for the selected hyperparameter:}
    \begin{itemize}
        \item Update the sub-arm weights for $J_t$:
        \[
        w_{I_t, j} \leftarrow w_{I_t, j} \cdot \exp\left(\frac{\gamma R_t}{Q_{t,j|I_t}}\right)
        \]
        \item Update the super-arm weights for $I_t$:
        \[
        w_{i} \leftarrow w_{i} \cdot \exp\left(\frac{\gamma R_t}{P_{t,i}}\right)
        \]
    \end{itemize}
\EndFor
\end{algorithmic}
\end{algorithm}

\section{Adversarial Nature of Hyperparameter Optimization in HABO}

\subsection{Evolving Landscapes and Adversarial Environments}

\hspace{2em}The HABO framework recognizes that hyperparameter tuning environments can exhibit adversarial characteristics. In a hierarchical structure like HABO—where hyperparameters are super-arms and their respective values are sub-arms—the reward from selecting a sub-arm can depend on earlier choices. For instance, setting \texttt{max\_features = sqrt} in a Random Forest may yield strong results initially but become less effective at generating accurate predictions when selected in another round.

\hspace{2em}This dynamic landscape demonstrates that a hyperparameter configurations reward changes over time, influenced by the optimizer’s own past actions, resembling an adversarial bandit setting.

\subsection{Psuedo-adversarial setting: Why EXP3?}

\hspace{2em}In hyperparameter tuning, revisiting a previously optimal value (e.g., {max\_features} of ``sqrt" in a Random Forest) may not yield the same results due to changes in other hyperparameters or model states. This creates a pseudo-adversarial environment where optimal configurations shift over time. As such, the problem of round-by-round modification can be viewed as adversarial.

\hspace{2em}Employing the EXP3 algorithm as the conduit of the HABO framework addresses this challenge by ensuring sublinear regret, specifically O(\(\sqrt{KT \log(K)}\)). This makes EXP3 well-suited for handling the dynamic, adversarial nature of hyperparameter tuning.

\section{Theoretical Insights}

\subsection{Insights on Discretized Values}
Discretized hyperparameter values offer several practical benefits:
\begin{itemize}
    \item \textbf{Efficient Exploration}: Limits the number of configurations to evaluate, reducing computational overhead.
    \item \textbf{Robustness to Noise}: Avoids overfitting to fine-grained parameter ranges, especially in noisy datasets.
    \item \textbf{Scalability}: Handles high-dimensional hyperparameter spaces effectively by focusing on meaningful ranges.
\end{itemize}

\subsection{Convergence Guarantees}

\hspace{2em}The convergence guarantees of HABO derive from its relationship to multi-armed bandit (MAB) algorithms, specifically EXP3 [10]. HABO inherits theoretical properties from EXP3, enabling sublinear regret against an adaptive adversary (which is the case). In essence, as the number of iterations $T$ grows large, the average regret per round approaches zero, and the algorithm allocates increasing probability mass toward the optimal arms. By modeling hyperparameters as ``super-arms'' and their potential values as ``sub-arms,'' HABO uses EXP3's proven convergence guarantees in a hierarchical setting.

\subsubsection{Foundations of Regret and Convergence in MAB}

Convergence in multi-armed bandits is commonly understood in terms of regret. Let there be $K$ arms and define the regret $R_T$ after $T$ rounds as
\[
R_T = \max_{i \in [k]} \sum_{t=1}^T x_{t,i} - \mathbb{E}\left[\sum_{t=1}^T x_{t,A_t}\right],
\]
where $x_{t,i}$ is the reward of arm $i$ at time $t$, and $A_t$ is the arm chosen by the algorithm at round $t$. Here, $x_{t,i}$ can be picked by an adversary, making the problem more challenging. 

A no-regret algorithm ensures that 
\[
\frac{R_T}{T} \to 0 \quad \text{as } T \to \infty,
\]
implying that the algorithm's performance asymptotically approaches that of the best fixed arm in hindsight. Thus, the probability of playing suboptimal arms diminishes over time, and the algorithm ``converges'' in the practical sense of increasingly selecting near-optimal actions.

\subsection{EXP3 Regret Bounds and Their Implications for HABO}

The EXP3 algorithm\cite{auer2001nonstochastic}  achieves a regret bound of
\[
R_T = O\left(\sqrt{T K \ln K}\right).
\]
This ensures that 
\[
\frac{R_T}{T} = O\bigg(\sqrt{\frac{K \ln K}{T}}\bigg) \to 0 \quad \text{as } T \to \infty.
\]

\vspace{1em}
\hspace{2em}Since HABO employs EXP3 (or an EXP3-like scheme with a hierarchical format) to pick both hyperparameters (super-arms) and their respective values (sub-arms), it inherits the no-regret guarantees. Each layer of HABO can be viewed as a separate bandit problem, and each bandit problem, observed by EXP3, achieves sub-linear regret. Thus, as $T \to \infty$, each component of the hyperparameter selection process zeroes in on near-optimal values.

\hspace{2em}Under these conditions, HABO converges to playing near-optimal configurations with high probability over time. The discretization of hyperparameter values means that HABO will identify the best choices within the chosen grid as $T$ increases.

\begin{figure}[htbp]
    \centering
    \includegraphics[width=0.7\linewidth]{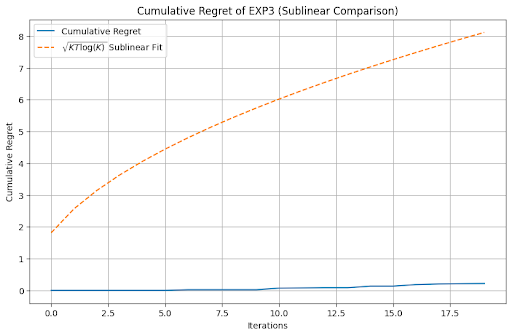}
    \caption{Python3 Habo Cumulative regret vs EXP3 upper bound\cite{karroum2024github}}
    \label{fig:enter-label}
\end{figure}

\subsection{Summary of Convergence Guarantees}

\hspace{2em}In summary, the convergence guarantees of HABO rest on well-established bandit theory. As $T$ grows large, HABO achieves sub-linear regret in selecting both which hyperparameter to tune and which value to assign. Consequently, the frequency with which HABO chooses near-optimal hyperparameter configurations increases, and the algorithm ``converges'' to solutions that are competitive with the best fixed configuration in hindsight.

\subsection{Comparison to Bayesian Optimization Regret Bounds}

Bayesian Optimization (BO) methods, such as the GP-UCB algorithm introduced by Srinivas\cite{Srinivas2012} and further explored by Desautels\cite{desautels2014bayesian}, achieve cumulative regret bounds of the form:
\[
R_T = O\left(\sqrt{T \beta_T \gamma_T}\right),
\]
Where $\beta_T$ comes from confidence interval parameters and $\gamma_T$ is the 
maximum mutual information that can be gained from $T$ observations. This BO model attempts to build a detailed process to guess how good a set of hyperparameters is. If the problem houses noisy or complex data, it causes the model to accumulate heavier computations, causing the regret to grow much quicker
\cite{desautels2014bayesian}.

Comparatively, the HABO framework leverages multi-armed bandit 
strategies (e.g., EXP3), which yield regret bounds on the order of 
\[
R_T = O\left(\sqrt{T K \ln K}\right),
\]
 
This bound does not explicitly depend on complex information-theoretic quantities 
like $\gamma_T$, offering a more straightforward control of regret as $T$ increases. By voiding the need for Gaussian Process models and their 
associated mutual information terms, HABO provides a scalable and robust alternative 
to BO methods that is less susceptible to issues like dimensionality and adversarial noise.

\section{Experimental Results and Discussion}

\subsection{Preface}
\hspace{2em}An implementation of the EXP3-based HyperArm Bandit Optimization (HABO) framework was developed using Python 3. It was applied to two datasets obtained from Kaggle to optimize hyperparameters and evaluate performance.

\hspace{2em}For classification tasks, the Titanic Dataset was utilized\cite{kaggle_titanic}, with model performance measured using accuracy. Accuracy quantifies the proportion of correct predictions out of the total predictions, providing a clear measure of the model’s ability to classify data correctly.

\hspace{2em}For regression tasks, the House Prices Dataset was used\cite{kaggle_houseprices}, with performance measured via the R-squared metric. R-squared (R2R2) represents the proportion of variance in the target variable that is explained by the model, giving insight into the model’s predictive accuracy in continuous output tasks.

\hspace{2em}To compare results, the same datasets were optimized using the scikit-learn Bayesian Optimization module in Python 3. The results demonstrated the computational efficiency and superior or comparable performance of the HABO framework against Bayesian Optimization, showcasing HABO’s strength in both classification and regression tasks. 

\subsection{Classification Task Case Study: Titanic Dataset}
\begin{itemize}
    \item \textbf{Case 1}: 20 rounds
    \item \textbf{Initial Accuracy}: 0.7762.
    \item \textbf{Best Accuracy via HABO (EXP3)}: 0.8112 in 6.36 seconds.
    \item \textbf{Best Accuracy via Bayesian Optimization}: 0.8042 in 10.75 seconds.

\item \textbf{Case 2}: 50 rounds
    \item \textbf{Initial Accuracy}: 0.7762.
    \item \textbf{Best Accuracy via HABO (EXP3)}: 0.8112 in 18.16 seconds.
    \item \textbf{Best Accuracy via Bayesian Optimization}: 0.8112 in 58.87 seconds.
\end{itemize}

\subsection{Regression Task Case Study: House Prices}
\begin{itemize}
    \item \textbf{Case 1}: 20 rounds
    \item \textbf{Initial Accuracy}: 0.8588.
    \item \textbf{Best Accuracy via HABO (EXP3)}: 0.8941 in 100.94 seconds.
    \item \textbf{Best Accuracy via Bayesian Optimization}: 0.8916 in 48.13 seconds.
\end{itemize}

\begin{itemize}
    \item \textbf{Case 1}: 50 rounds
    \item \textbf{Initial Accuracy}: 0.8588.
    \item \textbf{Best Accuracy via HABO (EXP3)}: 0.8953 in 194.92 seconds.
    \item \textbf{Best Accuracy via Bayesian Optimization}: 0.8945 in 203.50 seconds.
\end{itemize}

\begin{figure}[htbp]
    \centering
    \includegraphics[width=0.75\linewidth]{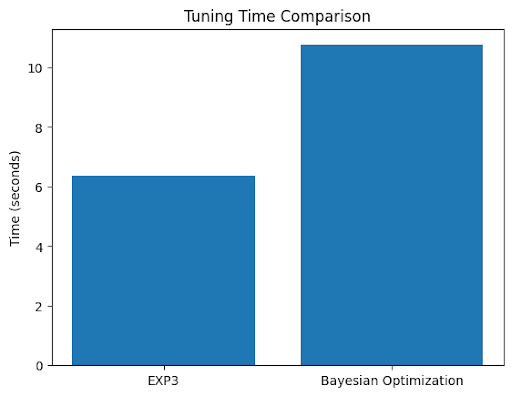}
    \caption{Titanic Tuning time bar graph 20 rounds: HABO vs BO\cite{karroum2024github}}
    \label{fig:enter-label}
\end{figure}

\begin{figure}
    \centering
    \includegraphics[width=0.75\linewidth]{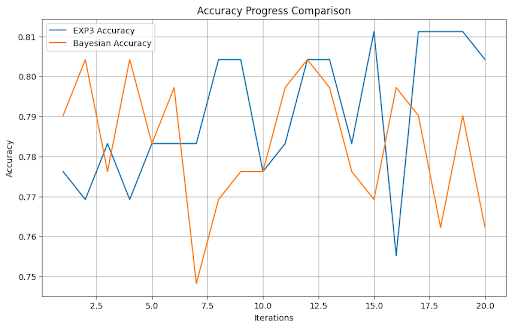}
    \caption{Titanic Accuracy progression over 20 rounds: HABO vs BO\cite{karroum2024github}}
    \label{fig:enter-label}
\end{figure}

\begin{figure}
    \centering
    \includegraphics[width=0.75\linewidth]{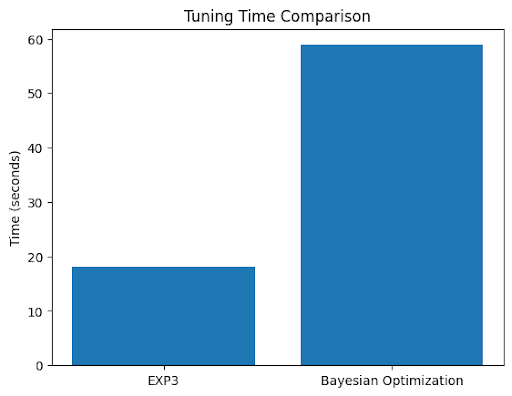}
    \caption{Titanic Tuning time bar graph 50 rounds: HABO vs BO\cite{karroum2024github}}
    \label{fig:enter-label}
\end{figure}

\begin{figure}
    \centering
    \includegraphics[width=0.75\linewidth]{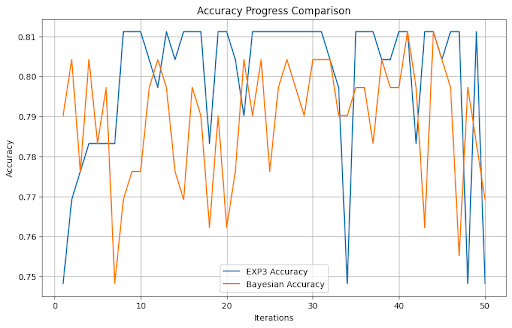}
    \caption{Titanic Accuracy progression over 50 rounds: HABO vs BO\cite{karroum2024github}}
    \label{fig:enter-label}
\end{figure}

\begin{figure}
    \centering
    \includegraphics[width=0.75\linewidth]{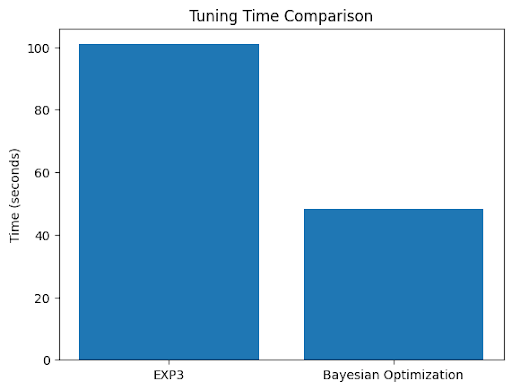}
    \caption{House Prices Tuning time bar graph 20 rounds: HABO vs BO\cite{karroum2024github}}
    \label{fig:enter-label}
\end{figure}

\begin{figure}
    \centering
    \includegraphics[width=0.75\linewidth]{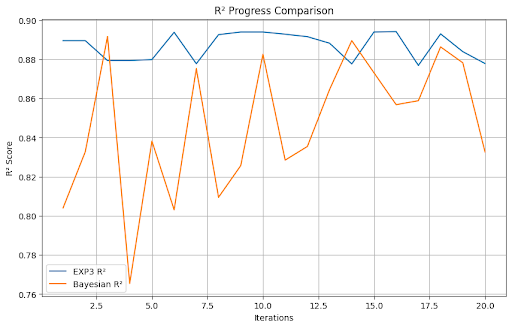}
    \caption{House Prices Accuracy progression over 20 rounds: HABO vs BO\cite{karroum2024github}}
    \label{fig:enter-label}
\end{figure}

\begin{figure}
    \centering
    \includegraphics[width=0.75\linewidth]{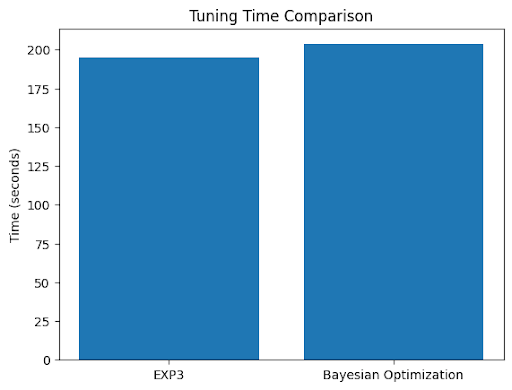}
    \caption{House Prices Tuning time bar graph 50 rounds: HABO vs BO \cite{karroum2024github}}
    \label{fig:enter-label}
\end{figure}

\begin{figure}
    \centering
    \includegraphics[width=0.75\linewidth]{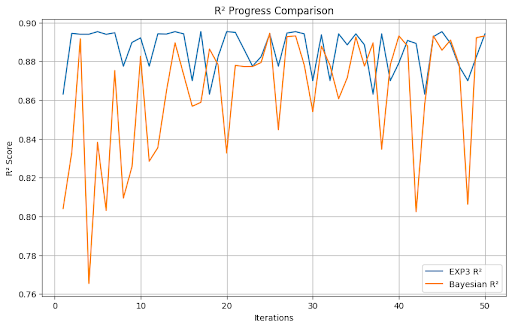}
    \caption{House Prices Accuracy progression over 50 rounds: HABO vs BO \cite{karroum2024github}}
    \label{fig:enter-label}
\end{figure}

\subsection{Insights}
\begin{itemize}
    \item HABO achieves faster convergence and higher accuracy/$R^2$ score compared to Bayesian Optimization.
    \item HABO compared to Bayesian Optimization performs exceptionally well in classification tasks in terms of performance and computational efficiency, however for regression tasks, falls short in computation but not in performance.
    \item Sub-arm granularity impacts computational cost and tuning precision.
    \item Robust performance in noisy settings demonstrates the strength of the adversarial bandit approach.
\end{itemize}

\section{Conclusion}

\hspace{2em}In this paper, we explored the theory and applications of bandit algorithms in both stochastic and adversarial settings, and introduced a novel framework, HABO, for hyperparameter optimization. We began by examining bandit algorithms—ETC, UCB, and EXP3—highlighting their theoretical guarantees, particularly in terms of regret analysis. While ETC and UCB rely on certain stochastic assumptions and provide robust bounds in those domains, EXP3 extends the analysis to adversarial environments, offering sublinear regret even in an adversarial setting.

\hspace{2em}Leveraging these insights, we developed HABO, which formulates hyperparameter optimization — an inherently complex and potentially adversarial problem—into a hierarchical bandit framework. By treating each hyperparameter as a super-arm and their possible values as sub-arms, HABO dynamically allocates computational effort to promising hyperparameter configurations. This adversarial bandit approach does not rely on Gaussian Process models or similar assumptions and therefore scales more efficiently, providing robustness in high-dimensional and noisy settings.

\hspace{2em}Our experimental results demonstrated that HABO can outperform or match the accuracy of Bayesian Optimization (BO) while often requiring less computational overhead.
Furthermore, due to its adversarial bandit foundation, HABO effectively handles evolving conditions where previously optimal hyperparameter values may cease to be the best possible configurations.
As the number of iterations increases, HABO’s sublinear regret guarantees ensure that it converges to near-optimal configurations over time.

\hspace{2em}In summary, HABO represents a new and powerful application of adversarial bandit algorithms to hyperparameter optimization. By bypassing the need for heavy modeling assumptions, HABO achieves a balance of scalability, efficiency, and theoretical soundness, providing a robust alternative to traditional tuning strategies.

\section{Future Work}

\hspace{2em}Future work with HABO could explore implementing different adversarial bandit algorithms, such as EXP4 or EXP3-IX, to determine which offers the best performance. Applications in fields like neural networks or other machine learning domains could also be investigated to assess the framework's potential in hyperparameter optimization. Additionally, comparisons with alternative hyperparameter optimization methods could provide valuable insights.


\begin{thebibliography}{99}

\bibitem{auer2002finite} 
P. Auer, N. Cesa-Bianchi, and P. Fischer. 
``Finite-time Analysis of the Multiarmed Bandit Problem." 
\textit{Machine Learning}, 47(2–3):235–256, 2002.

\bibitem{bubeck2012regret} 
S. Bubeck and N. Cesa-Bianchi. 
``Regret Analysis of Stochastic and Nonstochastic Multi-armed Bandit Problems." 
\textit{Foundations and Trends in Machine Learning}, 5(1):1–122, 2012.

\bibitem{lattimore2020bandit} 
T. Lattimore and C. Szepesvári. 
\textit{Bandit Algorithms.} 
Cambridge University Press, 2020. 
Available at: \url{https://banditalgs.com/}.

\bibitem{li2018hyperband}
L. Li, K. Jamieson, G. DeSalvo, A. Rostamizadeh, and A. Talwalkar.
``Hyperband: A Novel Bandit-Based Approach to Hyperparameter Optimization."
\textit{Journal of Machine Learning Research}, 18(185):1–52, 2018.

\bibitem{Srinivas2012} 
N. Srinivas, A. Krause, S. Kakade, and M. Seeger.
``Information-theoretic Regret Bounds for Gaussian Process Optimization in the Bandit Setting." 
\textit{IEEE Transactions on Information Theory}, 58(5):3250–3265, 2012.

\bibitem{desautels2014bayesian} 
T. Desautels, A. Krause, and J.W. Burdick.
``Parallelizing Exploration-Exploitation Tradeoffs in Gaussian Process Bandit Optimization." 
\textit{Journal of Machine Learning Research}, 15(1):3873–3923, 2014.

\bibitem{kaggle_titanic}
Kaggle. ``Titanic - Machine Learning from Disaster." 
\url{https://www.kaggle.com/c/titanic}

\bibitem{kaggle_houseprices}
Kaggle. ``House Prices - Advanced Regression Techniques."
\url{https://www.kaggle.com/c/house-prices-advanced-regression-techniques}

\bibitem{scikit-optimize}
T. Head, M. Kumar, H. Nahrstaedt, G. Louppe, and A. Gibiansky.
``Scikit-Optimize: Sequential model-based optimization in Python."
\url{https://scikit-optimize.github.io/stable/}

\bibitem{auer2001nonstochastic}
P. Auer, N. Cesa-Bianchi, Y. Freund, and R. Schapire. 
``The non-stochastic Multi-armed Bandit Problem." 
\textit{SIAM Journal on Computing}, 32(1):48–77, 2001. 
Available at: \url{https://citeseerx.ist.psu.edu/document?repid=rep1&type=pdf&doi=98db5a41d40f0d00f1328f9b0562ba027f9f0b2a}.

\bibitem{karroum2024github}
S. Karroum, S. Maazhar. ``Honours Project: Online Learning with Limited Feedback." 
GitHub repository, \url{https://github.com/simokarr/Honours-project-Online-learning-with-limited-feedback}

\bibitem{freund1995hedge}
Y. Freund and R. E. Schapire. 
``Game Theory, On-line Prediction and Boosting." 
In \textit{Proceedings of the Ninth Annual Conference on Computational Learning Theory (COLT)}, pp. 325–332, 1995.


\end{thebibliography}
\end{document}